\documentclass[conference]{IEEEtran}
\IEEEoverridecommandlockouts
\usepackage{cite}
\usepackage{amsmath,amssymb,amsfonts}
\usepackage{algorithmic}
\usepackage{graphicx}
\usepackage{textcomp}
\usepackage{xcolor}
\usepackage{csquotes}
\usepackage{siunitx}
\usepackage{subcaption}
\usepackage{gensymb}
\usepackage{url}
\usepackage{multirow}
\usepackage{colortbl}
\usepackage{hyperref}
\usepackage{cleveref}

\def\BibTeX{{\rm B\kern-.05em{\sc i\kern-.025em b}\kern-.08em
    T\kern-.1667em\lower.7ex\hbox{E}\kern-.125emX}}
\begin{document}

\title{Augmenting Off-the-Shelf Grippers with Tactile Sensing\\

\thanks{Remko Proesmans is a predoctoral fellow of the Research Foundation Flanders (FWO) under grant agreement no. 1S15923N. This work was also partially supported by the euROBIn Project (EU grant number 101070596).}
}

\author{\IEEEauthorblockN{1\textsuperscript{st} Remko Proesmans}
\IEEEauthorblockA{\textit{IDLab-AIRO} \\
\textit{Ghent University -- imec}\\
Ghent, Belgium \\
Remko.Proesmans@UGent.be}
\and
\IEEEauthorblockN{2\textsuperscript{nd} Francis wyffels}
\IEEEauthorblockA{\textit{IDLab-AIRO} \\
\textit{Ghent University -- imec}\\
Ghent, Belgium \\
Francis.wyffels@UGent.be}
}

\maketitle

\begin{abstract}
The development of tactile sensing and its fusion with computer vision is expected to enhance robotic systems in handling complex tasks like deformable object manipulation.
However, readily available industrial grippers typically lack tactile feedback, which has led researchers to develop and integrate their own tactile sensors. 
This has resulted in a wide range of sensor hardware, making it difficult to compare performance between different systems. 
We highlight the value of accessible open-source sensors and present a set of fingertips specifically designed for fine object manipulation, with readily interpretable data outputs. 
The fingertips are validated through two difficult tasks: cloth edge tracing and cable tracing.
Videos of these demonstrations, as well as design files and readout code can be found at
\href{https://github.com/RemkoPr/icra-2023-workshop-tactile-fingertips}{https://github.com/RemkoPr/icra-2023-workshop-tactile-fingertips}.

\end{abstract}


\section{Introduction}



Deformable object manipulation challenges several assumptions of classical robotics, like object rigidity, a low-dimensional state space and known dynamics models~\cite{zhu2022}.
Tactile sensing and its fusion with computer vision is expected to strengthen robotic systems in dealing with such complex deformable objects~\cite{billard2019}.
However, commercially available industrial grippers typically do not feature tactile feedback~\cite{borras2020, birglen2018}.
At most, binary object detection is included such that delicate parts can be grasped without exerting maximal gripping force.
This, combined with force torque sensing in the joints of a robotic arm, is the current extent of readily available tactile feedback.
Hence, researchers are required to integrate and even develop their own tactile sensors.
Sensor technologies in literature include piezoresistive~\cite{zlokapa2022}, capacitive~\cite{euan2022}, thermoelectric~\cite{li2020}, optoelectronic~\cite{cirillo2021} and more.
This heterogeneity in sensor hardware makes it difficult to compare the performance of control algorithms as the nature of the sensor is strongly intertwined with any control algorithm exploiting it.
Few dedicated robotic tactile sensor platforms have been adopted outside their founding research groups.
The emphasis of systems that do achieve external adoption, such as robot skins like uSkin~\cite{tomo2018} and tactile fingertips like GelSight~\cite{lambeta2020}, is transferability between robotic platforms.
These systems aim to be fit for a variety of robots and robotic grippers, so that researchers can acquire tactile add-ons to their pre-owned robotic infrastructure.
This hardware transferability facilitates task transferability, mitigating the issue of benchmarking in experiments making use of tactile sensing.
GelSight sensors, however, come with considerable overhead in terms of data processing and hardware integration, the latter being due to external cable routing.
We argue that there is plenty of room for accessible, open-source tactile fingertips with readily interpretable data outputs.
We present a set of narrow fingertips specifically suited for narrow object manipulation and demonstrate their potential in two actively explored problems in the state-of-the-art: cloth edge tracing~\cite{sunil2022} and cable tracing~\cite{cirillo2021, she2021}.

\section{Tactile Sensing for Narrow Objects} \label{ss:sensor_design}

\subsection{Electrical sensor design}
\begin{figure}[tpb]
  \centering
  \includegraphics[width=\linewidth]{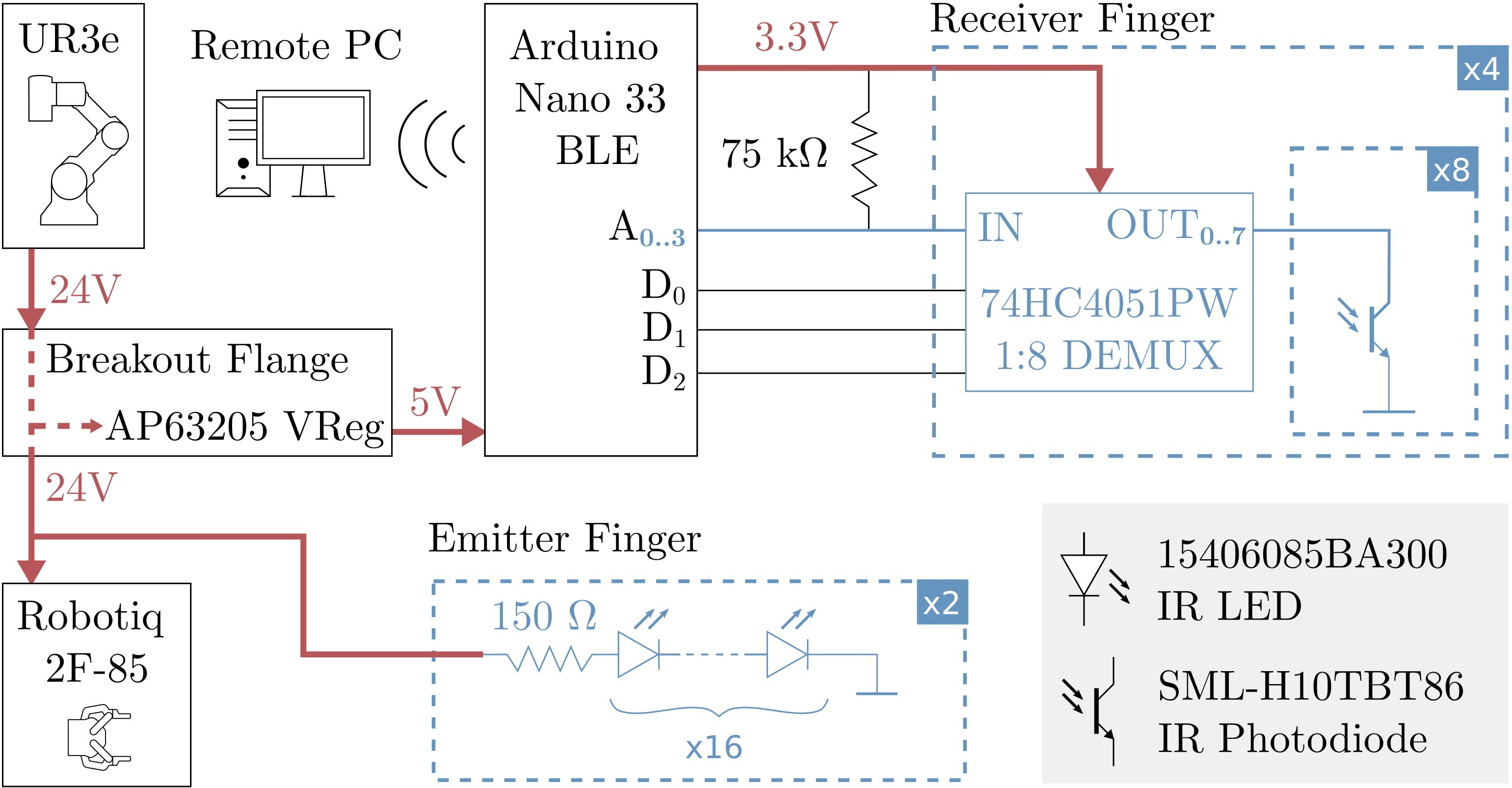}
  \caption{Electrical diagram of the tactile fingers integrated on a UR3e.}
  \label{fig:schematic}
\end{figure}

The tactile fingers are two complementary designs: one finger emits infrared (IR) light, the other receives it. 
The fingertips are integrated on a Robotiq~2F-85 gripper, mounted on a UR3e collaborative robot arm.
Fig.~\ref{fig:schematic} shows the schematic of the integrated system.
The first finger contains a grid of 32 IR light-emitting diodes (LEDs), connected in two parallel branches of each sixteen LEDs.
The grid is supplied directly by the \SI{24}{V} output of the UR3e arm via custom breakout flange between the Robotiq I/O Coupling and the gripper itself. 
Note that the I/O Coupling, and thus our sensor system, fits a multitude of different cobots and Robotiq grippers.
The LED finger consumes \SI{40}{\milli A} after a warmup time of five minutes.
The second finger consists of 32 IR photodiodes (PDs).
Here, four demultiplexers (DEMUX) are included, each addressing eight PDs.
The DEMUX are all simultaneously controlled by three digital pins of an Arduino Nano 33 BLE, such that they each route one of four analog pins to one of eight of their respective PDs.
In future work, a microcontroller will be integrated on the breakout flange, replacing the externally mounted Arduino.
The second finger consumes about \SI{1}{\milli A} of current from the \SI{3.3}{V} output of the Arduino, which in turn receives its power from the \SI{24}{V} UR3e tool output converted to \SI{5}{V} by a switching regulator on the breakout flange.
The discussed electrical specifications are summarised in Table~\ref{tab:tech_specs}.
The total current consumption reported includes the voltage regulator and the Arduino during readout.

\begin{table}[tpb]
\caption{Technical specifications of the tactile fingers.}
\label{tab:tech_specs}
\begin{center}
\begin{tabular}{ll}
\hline
\textbf{Characteristic} & \textbf{Value} \\
\hline
Total current consumption (incl. readout) & \SI{45}{\milli A} \\
\quad Receiver finger & \SI{1}{\milli A} \\
\quad Emitter finger & \SI{38}{\milli A} \\
Warmup time & 5\,min\\
\hline
Physical size (width $\times$ height $\times$ depth) & 58.0 $\times$ 54.0 $\times$ \SI{21}{\milli\metre\cubed} \\
\quad PCB dimensions & 42.8 $\times$ \SI{43.1}{\milli\metre\squared}  \\
\quad Sensorised area & \url{~}38.0 $\times$ \SI{30.0}{\milli\metre\squared} \\
\quad Thickness excluding gripper coupling & \SI{4.0}{\milli\metre} \\
Sensor density &\SI{3.21}{\per\centi\metre\squared} \\
\hline
Data resolution & 8\,bit \\
Readout frequency & \SI{55}{Hz} \\
\hline
\end{tabular}
\end{center}
\end{table}

\subsection{Structural sensor design \& manufacturing}

\begin{figure}[tpb]
  \centering
  \includegraphics[width=0.9\linewidth]{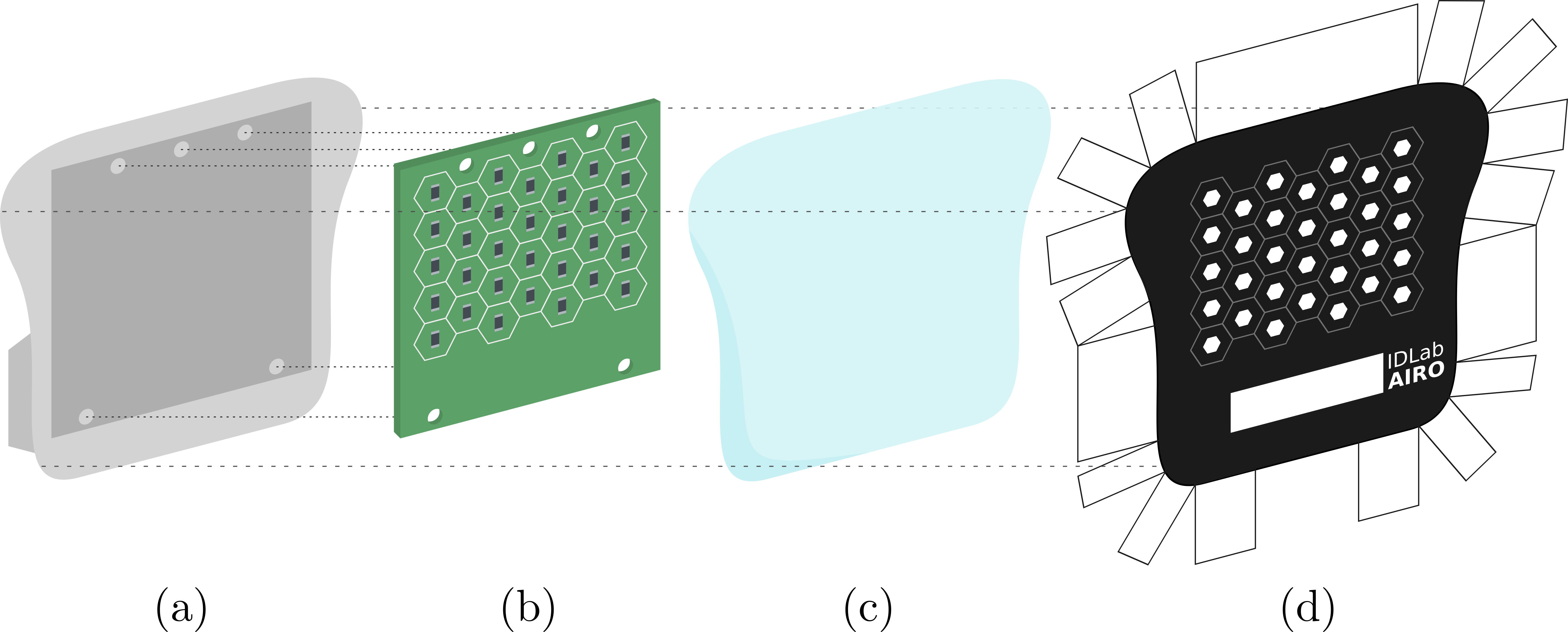}
  \caption{Sensor structure for both the emitting as well as the receiving finger. \textbf{(a)} 3D printed PETG backing. \textbf{(b)} PCB, either with 32 IR LEDs or 32 IR PDs. \textbf{(c)} \SI{1.5}{\milli\metre} thick layer of translucent silicone. \textbf{(d)} Sheet of \SI{100}{\micro\metre} Bemis 3914 TPU patterned with black ink.}
  \label{fig:sensor_structure}
\end{figure}

Structurally, both tactile fingers are largely similar.
Fig.~\ref{fig:sensor_structure} shows their constituent parts.
The IR LEDs and PDs are each placed in an identical hexagonal grid on their respective printed circuit boards (PCBs), such that a single LED is located straight across from a single PD when both fingers are brought close together.
For each finger, the PCB is slotted into a polyethylene terephthalate glycol (PETG) backing, 3D printed using a Prusa MK3 i3.
Alternatively, we have succeeded in producing a compliant version by printing the backing with Flexfill~TPU~98A and using a flexible PCB.  
The resulting structure is placed PCB side down into a mould filled with liquid Silicone Addition Colorless 5 obtained from Silicones and More.
After curing at room temperature for 24 hours, the finger is removed from the mould and a \SI{1}{mm} layer of translucent silicone protects the PCB components.
We have, however, observed degradation in the polymer right above the IR LEDs after \url{~}100 grasp cycles, so further material exploration is warranted.
However, the surface of this silicone layer is sticky and easily accumulates dirt.
Hence, a sheet of \SI{100}{\micro\metre} thick Bemis 3914 thermoplastic polyurethane (TPU) is shaped around the combined PETG, PCB and silicone structure by pulling the TPU taut and heating it locally using a CIF~852 hot air gun set to \SI{90}{\degree C}.
The TPU was found to provide a low-friction surface for cloth manipulation and is highly durable during normal use.
Additionally, before shaping, a hole pattern is printed in black on the inside of the TPU layer using an HP~LaserJet~P2015. 
Each hole is positioned directly above an IR emitter or receiver, thus focusing their respective radiation patterns.
The entire fingertip is \SI{4}{\milli\metre} thick and has a slanted tip, making it suitable for sliding underneath cloth for grasping, as opposed to e.g. GelSight sensors~\cite{lambeta2020}.
Table~\ref{tab:tech_specs} provides additional structural specifications.

\subsection{Sensor readout}

Getting a single readout from the tactile fingers entails sequentially setting the common selection inputs for the DEMUX to one of eight codes and reading the voltage on the four analog pins that are routed along the receiver grid.
This data is transmitted in an 8-bit format at about \SI{55}{Hz} by the Arduino over Bluetooth Low Energy (BLE).
Wireless communication avoids cable routing and thus reduces integration overhead. The use of a widespread communication protocol further encourages transferability.

When the fingers are close together, the PDs receive high IR intensity and pull down the analog pins to ground.
Grasped objects between the fingers block a large part of the IR light from the LEDs, which leads to the PDs conducting less and results in higher voltages read by the analog pins. 
This sensing principle is specifically suited towards, but not limited to, detecting exceedingly narrow objects. 
Thin cloth, for example, is tough to detect using traditional force sensors.
In the following, the 8-bit complement of the readings is reported, as it is more intuitive that high IR light transmission corresponds to high sensor readings. 

\section{Experimental validation}
\subsection{Response to different materials}
\begin{figure}[tpb]
  \centering
  \includegraphics[width=0.75\linewidth]{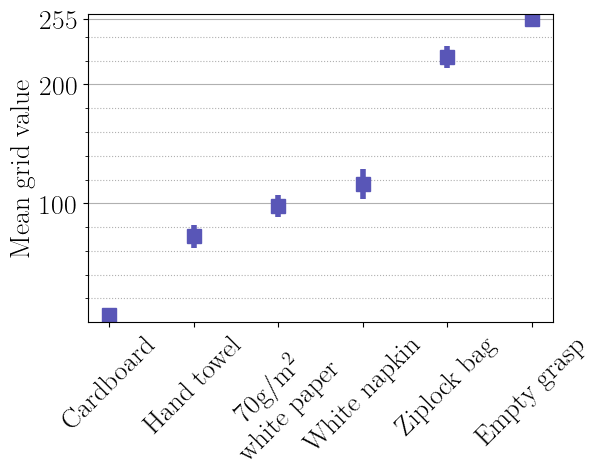}
  \caption{Mean sensor values when grasping different materials.}
  \label{fig:grasping_exp}
\end{figure}

The sensor readout depends on the translucency of the material that is grasped.
Fig.~\ref{fig:grasping_exp} shows the average 8-bit value received by the PD grid when placing different materials between the fingertips at a separation distance of \SI{6}{\milli\metre}.
The materials are solid cardboard, a pink hand towel, a sheet of white paper, a thin white cloth napkin and a clear plastic ziplock bag.
The results are intuitively clear: more translucent materials lead to higher sensor readings. 

\subsection{Task solving}
\begin{figure}[tpb]
\centering
\hspace{4.3cm}
\begin{subfigure}[t]{0.2\textwidth}
    \centering
    \includegraphics[height=1.2cm]{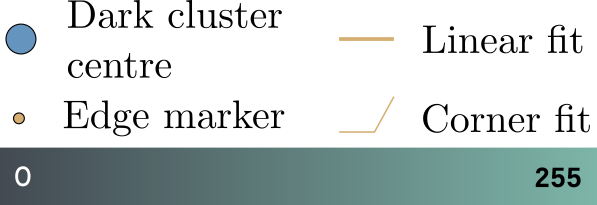}
\end{subfigure}\hfill\vspace{2mm}
\begin{subfigure}[t]{0.2\textwidth}
    \centering
    \includegraphics[height=3.45cm, trim={4cm 0cm 2cm 4cm}, clip]{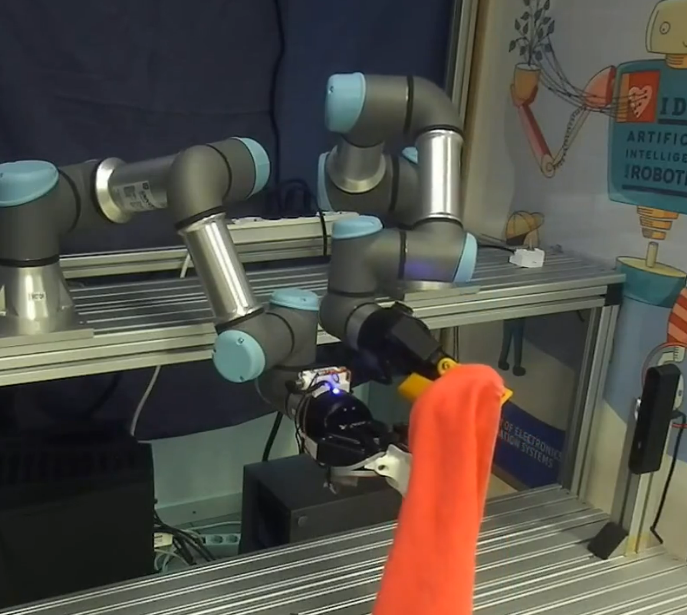}
    \caption{Grasping cloth edge.}
    \label{fig:exp_cloth_start_pic}
\end{subfigure}\hfill\vspace{2mm}
\begin{subfigure}[t]{0.25\textwidth}
    \centering
    \includegraphics[height=3.45cm, trim={1.5cm 3cm 0.5cm 1cm}, clip]{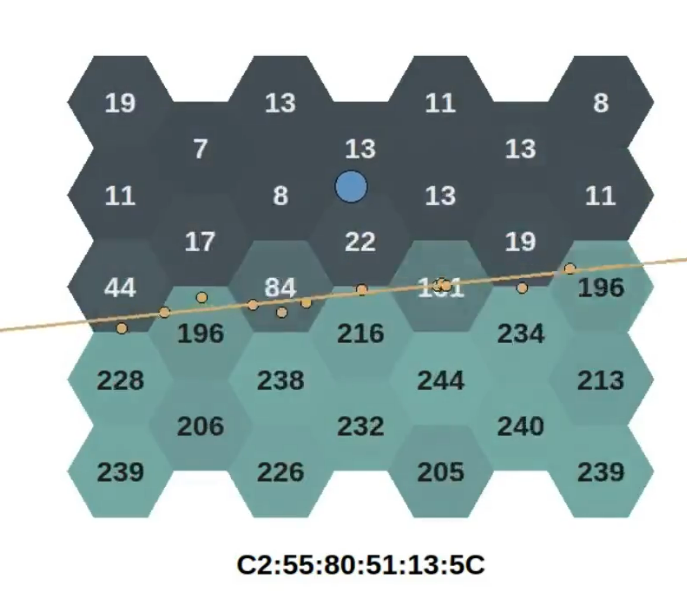}
    \caption{Sensor values corresponding to Fig.~\ref{fig:exp_cloth_start_pic}.}
    \label{fig:exp_cloth_start_grid}
\end{subfigure}\hfill\vspace{2mm}
\begin{subfigure}[t]{0.2\textwidth}
    \centering
    \includegraphics[height=3.45cm, trim={4cm 0cm 2cm 4cm}, clip]{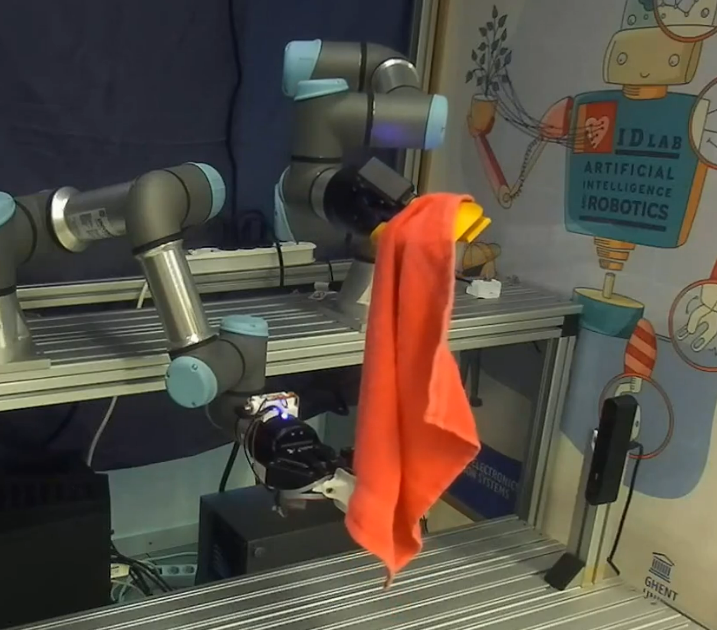}
    \caption{Grasping cloth corner.}
    \label{fig:exp_cloth_end_pic}
\end{subfigure}\hfill\vspace{2mm}
\begin{subfigure}[t]{0.25\textwidth}
    \centering
    \includegraphics[height=3.45cm, trim={1.5cm 3.5cm 0.5cm 0}, clip]{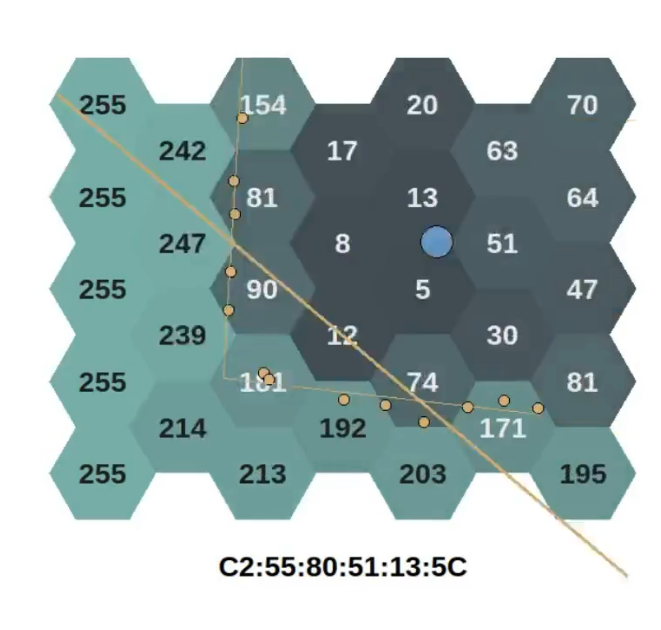}
    \caption{Sensor values corresponding to Fig.~\ref{fig:exp_cloth_end_pic}.}
    \label{fig:exp_cloth_end_grid}
\end{subfigure}\hfill\vspace{2mm}
\begin{subfigure}[t]{0.2\textwidth}
    \centering
    \includegraphics[height=3.45cm, trim={4cm 3cm 2cm 1cm}, clip]{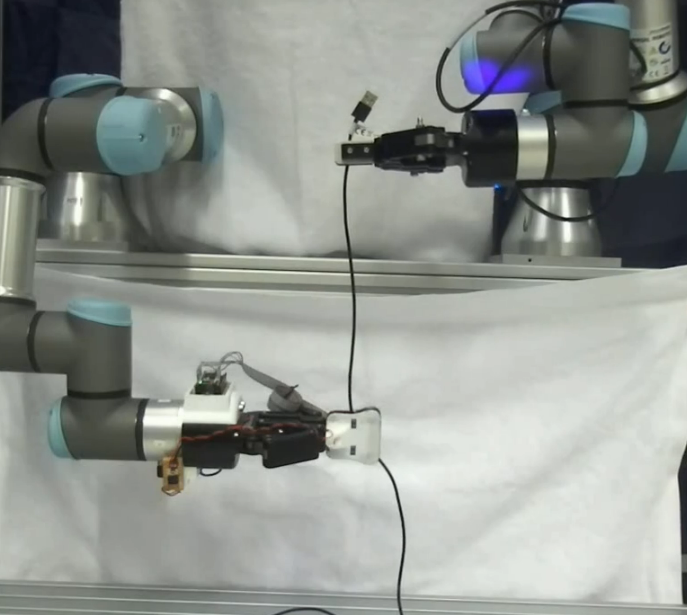}
    \caption{Grasping \SI{3}{\milli\metre} USB cable.}
    \label{fig:exp_cable_end_pic}
\end{subfigure}\hfill\vspace{2mm}
\begin{subfigure}[t]{0.25\textwidth}
    \centering
    \includegraphics[height=3.45cm, trim={0.5cm 3.5cm 0.5cm 0}, clip]{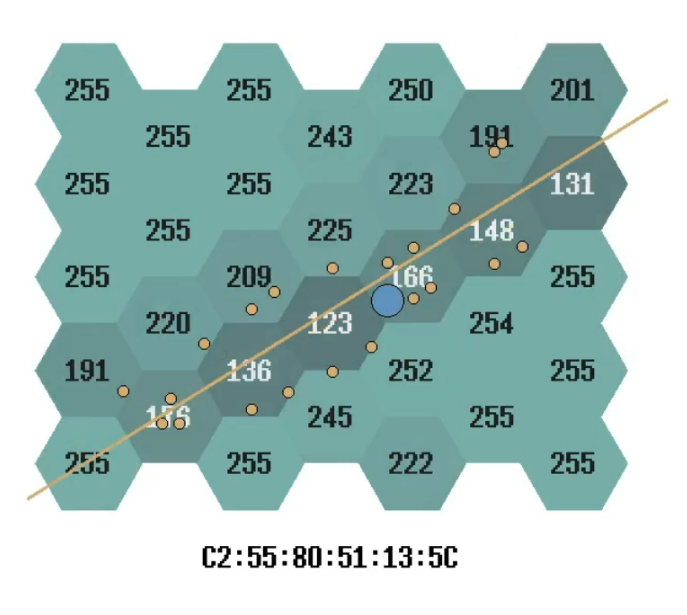}
    \caption{Sensor values corresponding to Fig.~\ref{fig:exp_cable_end_pic}.}
    \label{fig:exp_cable_end_grid}
\end{subfigure}\hfill\vspace{2mm}
\caption{Cloth and cable tracing.}
\label{fig:exp}
\end{figure}

The sensors are validated in two actively explored problems in the state-of-the-art: cloth edge tracing (Fig.~\ref{fig:exp_cloth_start_pic}-\ref{fig:exp_cloth_end_grid}) and cable tracing (Fig.~\ref{fig:exp_cable_end_pic}-\ref{fig:exp_cable_end_grid}).
The used methods are directly transferable to other robotic installations to which the fingertips can be fitted.
We refer the reader to the accompanying GitHub page for video material.

The tactile data is distilled into a set of control parameters as follows.
Each data value corresponds to a hexagonal cell, of which the position in the grid is defined by its centre coordinates.
In most use cases, the grid will show distinct dark and bright zones, see the visualisations in Fig.~\ref{fig:exp}.
Several processing steps are applied to the sensor data:
\begin{itemize}
    \item The threshold $\lambda$ deciding which values are \enquote{bright}, respectively \enquote{dark}, is calculated by sorting all grid values from lowest to highest and finding the largest step between adjacent values.
    Any values found before this step are \enquote{dark}, the others \enquote{bright}.
    $\lambda$ is then set to the average of the bright cluster mean and the dark cluster mean value.
    \item The centre point of the dark cluster is calculated as the weighted mean of the centre coordinates of all cells belonging to the dark cluster, where the weights are the 8-bit complements of the associated values.
    \item The shape of the border between the dark and the bright cluster indicates whether or not a corner is grasped when tracing cloth, or shows the angle at which the cable is passing through the fingers.  
    To find the border shape, the value of each cell is compared to those of the neighbouring cells.
    If $\lambda$ is crossed between cells, the values of the cells are linearly interpolated along the line segment connecting the cell centre points and an edge marker is placed where an interpolated value of $\lambda$ is found.
    Piecewise linear functions are then fitted to these edge markers.
\end{itemize}
In both experiments, proportional control with manually tuned gain is used to maintain the centre point of the dark cluster at a certain height in the sensor grid.

\section{Conclusion}
Transferability in robotics is strongly hardware-limited.
In the field of tactile sensing, a mitigating approach is to design tactile fingertips, facilitating transfer of such sensors between heterogeneous robotic hardware and potentially allowing for direct control policy transferability.
We have presented a set of open-source optoelectronic tactile fingertips and have demonstrated their potential in cable and cloth edge tracing. 
The tactile fingertips are uniquely suited towards manipulation of narrow objects, while being thinner than e.g. GelSight sensors and the optoelectronic sensors in \cite{cirillo2021}, allowing for a compliant design.
Additionally, sensor platforms like ours are more accessible than GelSight sensors in terms of cost and both hardware and software integration, further incentivising benchmarking in tactile robotics.

\end{document}